\documentclass{article}
\usepackage{ijcai21}

\usepackage{times}

\usepackage{soul}
\usepackage{url}
\usepackage[hidelinks]{hyperref}
\usepackage[utf8]{inputenc}
\usepackage[small]{caption}
\usepackage{graphicx}
\usepackage{amsmath}

\usepackage{multicol}
\usepackage{multirow}
\usepackage{diagbox}
\usepackage{booktabs}
\usepackage{subcaption}
\usepackage{bbold}

\usepackage{amssymb}
\usepackage[ruled, linesnumbered]{algorithm2e}
\usepackage{xcolor}
\usepackage[symbol]{footmisc}
\usepackage{color}

\usepackage{algpseudocode}

\title{Contrastive Model Inversion for Data-Free Knowledge Distillation}

\author{
Gongfan Fang$^{1,3}$
\and
Jie Song$^{1}$\footnote{Corresponding author.}\and
Xinchao Wang$^{2}$\and
Chengchao Shen$^1$\and \\
Xingen Wang$^1$\and 
Mingli Song$^{1,3}$
\affiliations
$^1$Zhejiang University\and
$^2$National University of Singapore\\
$^3$Alibaba-Zhejiang University Joint Research Institute of Frontier Technologies\\
\emails
\{fgf, sjie,chengchaoshen,newroot,brooksong\}@zju.edu.cn,
xinchao@nus.edu.sg,
}

\begin{document}

\maketitle

\newcommand{\etal}{\textit{et al}. }

\begin{abstract}
   Model inversion, whose goal is to recover training data from a pre-trained model, has been recently proved feasible. However, existing inversion methods usually suffer from the mode collapse problem, where the synthesized instances are highly similar to each other and thus show limited effectiveness for downstream tasks, such as knowledge distillation. In this paper, we propose Contrastive Model Inversion~(CMI), where the data diversity is explicitly modeled as an optimizable objective, to alleviate the mode collapse issue. Our main observation is that, under the constraint of the same amount of data, higher data diversity usually indicates stronger instance discrimination. To this end, we introduce in CMI a contrastive learning objective that encourages the synthesizing instances to be distinguishable from the already synthesized ones in previous batches. Experiments of pre-trained models on CIFAR-10, CIFAR-100, and Tiny-ImageNet demonstrate that CMI not only generates more visually plausible instances than the state of the arts, but also achieves significantly superior performance when the generated data are used for knowledge distillation. Code is available at \url{https://github.com/zju-vipa/DataFree}.
\end{abstract}

\section{Introduction}
Recent advances in deep learning have lead to a vast number of publicly available pre-trained models, which covers a variety of scenarios such as recognition~\cite{krizhevsky2012imagenet} and detection~\cite{ren2015faster}. Albeit their striking performance, most of these pre-trained models are cumbersome in size and problematic to be deployed into capacity-sensitive edge devices. To resolve this issue, knowledge distillation (KD) was proposed to craft a lower-latency model by learning dark knowledge from pre-trained teacher models. It soon became a flourishing topic due to its effectiveness and engineering simplicity,
and has been adopted to both Euclidean and non-Euclidean data~\cite{Hinton2015DistillingTK,yang2021distilling}

Despite the promising results achieved, existing KD methods largely rely on massive training data to transfer the knowledge from pre-trained teacher models to students. However, in many cases, the training data is not released together with the pre-trained models, due to privacy or transmission reasons, which makes these methods inapplicable. Data-free KD is thus proposed to address this problem. The vital step of data-free KD is model inversion, for which the goal is to recover training data from pre-trained teacher models. With the derived synthesized data, student models can be learned with ease through directly leveraging the data-driven KD methods. 

Model inversion \emph{per se} has been studied for a long time for various purposes. For example,~\cite{mahendran2015understanding} investigated model inversion for better understanding the deep representations.~\cite{Fredrikson2015ModelIA} studied model inversion attack to infer sensitive information. Recently, the study of model inversion resurges as data-free KD gains more attention~\cite{yin2020dreaming,lopes2017data,fang2019data}. Specifically, data-free KD imposes higher demands on model inversion, due to the following reasons. First, the derived data should follow the same distribution as the original training data, since otherwise the student model can barely inherit knowledge from the teacher with these out-of-distribution data. Second, the derived data should be of rich diversity, so that the student model is able to master the comprehensive knowledge with these diverse data.

Existing inversion methods, unfortunately, remain incompetent in meeting such demands. The work~\cite{chen2019data}, for instance, inverted the classification model by fitting the ``one-hot'' prediction distribution.~\cite{yin2020dreaming}, on the other hand, synthesized images by regularizing the distribution of intermediate feature maps using statistics stored in batch normalization layers of the teacher model. Both methods rely on some assumptions on the true data distribution and optimize each instance independently by fitting the prior distribution. As no constraints are explicitly imposed to encourage data diversity, these methods suffer from the mode collapse problem, where the generated instances turn out to be highly similar to each other. The works of~\cite{fang2019data,choi2020data} proposed to generate more discriminant data for training by mining harder or adversarial examples.  Although some performance gain is achieved for data-free KD, the generated data often lack natural image statistics and are easily identified as unrealistic.

In this paper, we attempt to alleviate the mode collapse problem in data-free KD, through the lens of promoting data diversity. From experiments, we observe that with the same amount of data, higher data diversity indicates stronger instance discrimination. Inspired by this phenomenon, we first propose a definition of data diversity based on instance discrimination, and then propose the Contrastive Model Inversion~(CMI) for tackling mode collapse while keeping the distribution of generated data closer to real data distribution. In this way, the generated data becomes more diverse and realistic.

Specifically, in CMI we introduce another contrastive learning objective, where the positive image pair comprises a cropped image and whole image of the same data instances, while the negative image pair comprises two different data instances. By encouraging positive pairs of images to be close to each other and negative pairs of images to be isolated apart, CMI significantly improves the image diversity and realism, and therefore promotes the performance of data-free KD. Experiments of pre-trained models on CIFAR-10, CIFAR-100, and Tiny-ImageNet demonstrate that CMI not only ensures more visually plausible instances than the state of the arts, but also achieves significantly superior performance when the generated data are used for knowledge distillation.

Our contributions are therefore summarized as follows.
\begin{itemize}
\item We propose a definition of data diversity, which enables us to incorporate the diversity into the optimization objective explicitly to improve the diversity of generated data.
\item We propose a novel Contrastive Model Inversion approach to handle
mode collapse in data-free KD while
enforcing the distribution of generated data more
to imitate the real one. 
\item Extensive experiments are conducted to validate 
the superiority of CMI with respect to the state of the arts.
\end{itemize}

\section{Related Works}

\paragraph{Model Inversion (MI)} aims at re-constructing inputs from the parameter of a pre-trained model, which is originally proposed to understand the deep representation of neural networks~\cite{mahendran2015understanding}. Given a function mapping $\phi(x)$ and the input $x$, a standard model inversion problem can be formalized as finding a $x^{\prime}$ to achieve the smallest $d(\phi(x), \phi(x^{\prime}))$, where $d(\cdot, \cdot)$ is a error function, e.g., mean squared error. This paradigm, known as model inversion attack~\cite{wu2016methodology}, is widely used in several areas such as model security~\cite{zhang2020secret} and interpretability~\cite{mahendran2015understanding}. Recently, the inversion techniques have shown their effectiveness in knowledge transfer~\cite{lopes2017data,yin2020dreaming}, leading to data-free knowledge distillation.

\paragraph{Data-Free Knowledge Distillation}
recently emerges as a popular task within
the realm of knowledge distillation~\cite{YeCVPR2019,ShenAAAI19,yang2020factorizable,YeCVPR20},
and aims to learn a student model from a cumbersome teacher without accessing real-world data~\cite{lopes2017data,chen2019data,ma2020adversarial}, so as to achieve model compression~\cite{YuCVPR17}. The contributions of existing data-free works can be roughly divided into two categories: adversarial training and data prior. Adversarial training is motivated by robust optimization, where worst-case samples are synthesized for student learning~\cite{micaelli2019zero,fang2019data}. Data prior provide another perspective for data-free KD, where the synthetic data are forced to satisfy a pre-defined prior, such as total variance prior~\cite{mahendran2015understanding} and Batch normalization statistics~\cite{yin2020dreaming}. 

\paragraph{Contrastive Learning} has achieved tremendous progress in field of self-supervised learning~\cite{chen2020simple,he2020momentum}. Its core idea to treat each sample as a distinct category and learn how to distinguish them~\cite{wu2018unsupervised,LiuAAAI21}. In this work, we re-visit the contrastive learning framework from another perspective, where its ability of instance discrimination is used to model the data diversity in model inversion. 

\begin{figure*}[t]
   \begin{center}
      \includegraphics[width=0.95\linewidth]{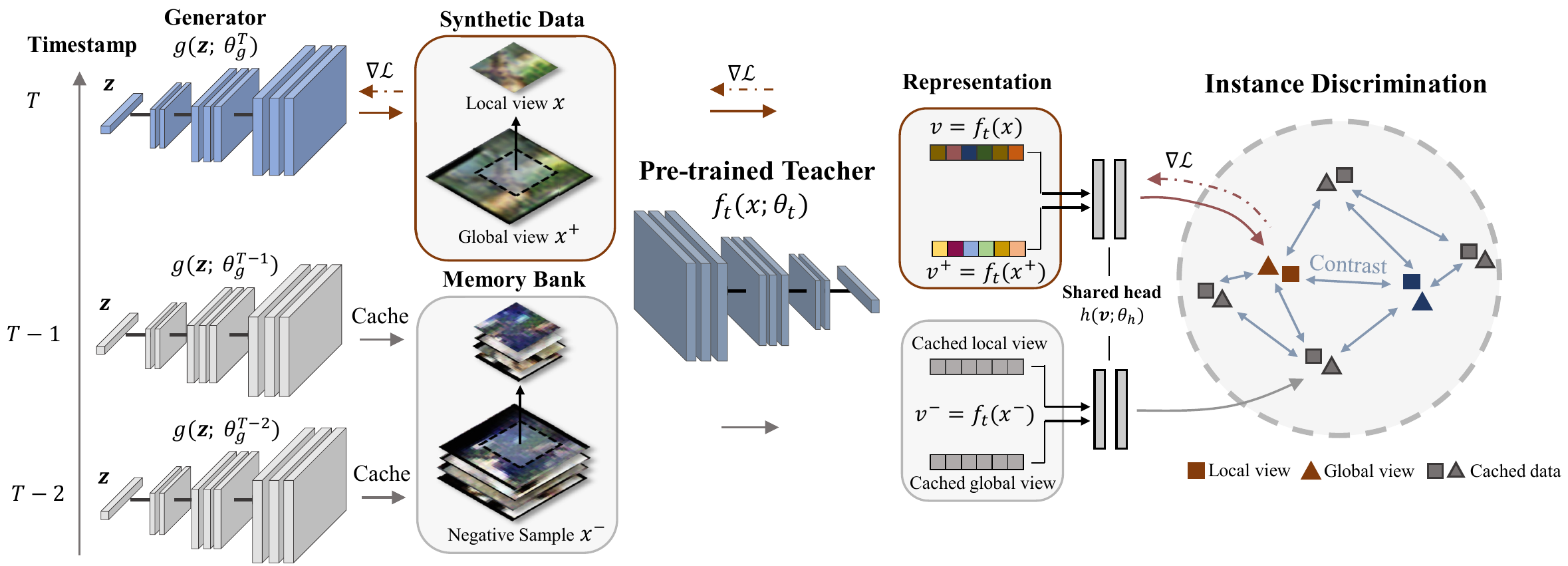}
   \end{center}
         \vspace{-1.2em}
      \caption{The illustrative diagram of proposed contrastive model inversion approach. In each timestamp, a freshly initialized generator is trained to synthesize distinguishable samples under the guidance of instance discrimination.} \label{fig:framework}
      \vspace{-1.5em}
\end{figure*}

\section{Method}


\subsection{Preliminary} \label{sec:preliminary}

Model inversion, as a vital step for data-free knowledge distillation, aims to recover training data $\mathcal{X^{\prime}}$ from a pre-trained teacher model $f_t(x; \theta_t)$ as an alternative to the inaccessible original data $\mathcal{X}$. In this part, we discuss three typical inversion techniques:

\paragraph{BN regularization} is originally introduced in ~\cite{yin2020dreaming} to regularize the distribution of $\mathcal{X}$ by making an Gaussian assumption about features from network. The regularization is usually represented as the divergence between the feature statistics $\mathcal{N}(\mu_l(x), \sigma_l^2(x))$ and Batch normalization statistics $\mathcal{N}(\mu_{l}, \sigma^2_{l})$ as follows:
\begin{equation}
   \mathcal{L}_{bn}(x) = \sum_{l} D\left( \mathcal{N}(\mu_l(x), \sigma_l^2(x)), \mathcal{N}(\mu_{l}, \sigma^2_{l}) \right)
\end{equation}

\paragraph{Class prior} is usually introduced for class-conditional generation, which makes an ``one-hot'' assumption about the network predictions on $x\in \mathcal{X^{\prime}}$~\cite{chen2019data}. Given a pre-defined category $c$, it encourages to minimize the cross entropy loss: 
\begin{equation}
   \mathcal{L}_{cls}(x) = CE(f_t(x), c)
\end{equation}

\paragraph{Adversarial Distillation} is motivated by robust optimization, where the $x$ is forced to produce large disagreement between teacher $f_t(x; \theta_t)$ and student $f_s(x; \theta_s)$~\cite{micaelli2019zero,fang2019data}, i.e., maximize a Kullback–Leibler divergence term:
\begin{equation}
   \mathcal{L}_{adv}(x) = - KL(f_t(x) / \tau \| f_s(x) / \tau)\label{eqn:adv}
\end{equation}

\paragraph{Unified Framework} Combining the above mentioned techniques will lead to a unified inversion framework~\cite{choi2020data} for data-free knowledge distillation:
\begin{equation}
   \mathcal{L}_{inv} = \alpha \cdot \mathcal{L}_{bn}(x) + \beta \cdot \mathcal{L}_{cls}(x) + \gamma \cdot \mathcal{L}_{adv}(x) \label{eqn:unified}
\end{equation}
where $\alpha$, $\beta$ and $\gamma$ are balance terms for different criteria. However, model inversion is actually a ill-conditioned problem where the synthetic data satisfying the above criteria are not unique. As no diversity constraints are explicitly imposed in this framework, the conventional inversion method may tend to be ``lazy'' and repeatedly synthesize duplicated samples. To overcome this problem, we propose a diversity-aware inversion technique, namely Contrastive Model Inversion (CMI).

\subsection{Contrastive Model Inversion}

\paragraph{Overview} With a pre-trained teacher model $f_t(x; \theta_t)$, the goal of contrastive model inversion is to produce a set of $x\in \mathcal{X^{\prime}}$ with strong diversity, with which comprehensive knowledge can be extracted from teachers. In this section, we develop an interesting definition for data diversity and, based on this, introduce the proposed Contrastive Model Inversion (CMI). Our motivation is intuitive: under the constraint of the same data amount, higher diversity usually indicates stronger instance distinguishability. To this end, we model the data diversity with an instance discrimination problem~\cite{wu2018unsupervised} and construct an optimizable objective with contrastive learning.

\paragraph{Definition of Data Diversity} Given a set of data $\mathcal{X^{\prime}}$, An intuitive description of data diversity would be ``how distinguishable are the samples from the dataset'', which reveals a positive correlation between the diversity and instance distinguishability. Thus if we have a certain metric $d(x_1, x_2)$ to estimate the distinguishability for an instance pair $\{x_1, x_2\}$, then we can develop a clear definition for data diversity as the following:
\begin{equation}
   \mathcal{L}_{div}(\mathcal{X}) = \mathbb{E}_{x_1, x_2 \in \mathcal{X}} \left[ d(x_1, x_2) \right] \label{eqn:div}
\end{equation} 
where $d(x_1, x_2)$ will be applied to all possible $(x_1, x_2)$ pairs from $\mathcal{X}$. There are various ways to define the metric $d(\cdot, \cdot)$, leading to different diversity criteria. For example, the pre-trained model $f_t(x; \theta_t)$ is actually an embedding function that maps data $x$ to a high-level feature space, where a naive metric can be defined as $d(x_1, x_2)=\|f_t(x_1)-f_t(x_2)\|$, which is known as perception distance~\cite{li2003discovery}. However, this distance may be problematic for diversity estimation due to the following issues: 1) the function $f_t$ is actually not explicitly trained for measuring the similarity between instances, where the meaning of euclidean distance is unknown to us. 2), the embedding $f_t(x)$ may encode structured information about inputs, which can not be captured by this metric. 3) this distance metric is unbounded, and we are not to figure out how large it should be to indicate a good diversity. In this case, maximizing such a distance metric on $f_t$ may only lead to abnormal results. Thus, a more appropriate embedding space are required to construct a meaningful metric for distinguishability. In what follows, we propose a learning-based metric for data diversity, which is established by solving a contrastive learning objective. 

\paragraph{Data Diversity from Contrastive Learning} \label{sec:diversity}  contrastive learning is original proposed to learn useful representations from data in a self-supervised manner, where an instance-level discrimination problem is established by treating each sample as a distinct category~\cite{wu2018unsupervised}. Through contrastive learning, a network can learn how to distinguish different instances, which exactly coincides with our requirements for the metric $d(\cdot, \cdot)$. Based on this, we introduce another network $h(\cdot)$ as an instance discriminator upon the teacher network $f_t$ that accepts feature $f_t(x)$ as input and project it into a new embedding space. For simplification, we use $v=h(x)$ to represent $v=(h \circ f_t)(x)$ because the teacher network are fixed. In the new embedding space of $h(\cdot)$, we use simple cosine similarity to describe the relationship between data pair $x_1$ and $x_2$ as the following:
\begin{equation}
   sim(x_1, x_2, h) = \frac{ \langle h(x_1), h(x_2) \rangle }{\|h(x_1)\|\cdot\|h(x_2)\|}
\end{equation} 

Then, an instance discrimination problem can be formulated in the form of contrastive learning framework~\cite{chen2020simple}, where each instance will be randomly transformed into different views and should be matched correctly. For each instance $x\in \mathcal{X^{\prime}}$, we construct a positive view $x^+$ by random augmentation and treat other instances $x^-$ as negative ones. The contrastive learning loss is formalized as follows:
\begin{equation}
   \mathcal{L}_{cr}(\mathcal{X}, h) = - \mathbb{E}_{x_i \in \mathcal{X}} \left[ log \frac{exp(sim(x_i, x^+_i, h)/\tau)}{\sum_j exp(sim(x_i, x^-_j, h)/\tau)} \right] \label{eqn:cr}
\end{equation}
where $\tau$ denotes the temperature. Discriminator $h(\cdot)$ can learn how to distinguish different samples by push positive pairs closer and pull negative pairs apart, which provide a ``contrast'' metric for any $(x_1, x_2)$ pair. To further understand how it related to the data diversity, let's define the metric $d(x_1, x_2)$=$-\frac{exp(sim(x_1, x_2, h)/\tau)}{exp(sim(x_1, {x_1}^+, h)/\tau)}$ and rewrite Eqn. \ref{eqn:div}: 
\begin{equation}
   \begin{split}
   \mathcal{L}_{div}(\mathcal{X}) & = -\mathbb{E}_{x_i\in \mathcal{X}} \mathbb{E}_{x_j \in \mathcal{X}} \left[ \frac{ exp(sim(x_i, {x_j}^-, h)/\tau) }{ exp(sim(x_i, {x_i}^+, h)/\tau) }  \right] \\
   & = -\frac{1}{Z(x^-)}\mathbb{E}_{x_i\in \mathcal{X}} \left[  \frac{\sum_{j} exp(sim(x_i, x^-_j, h)/\tau)}{exp(sim(x_i, x^+_i, h)/\tau)} \right] \\
   & = \frac{1}{Z(x^-) \cdot \mathcal{L}_{cr}(\mathcal{X}, h) }
   \end{split} \label{eqn:cr-div}
\end{equation} 
where the constant $Z(x^-)$ refers to the amount of negative samples for each instance $x_i$. Thus we can directly maximize the diversity $\mathcal{L}_{div}$ by minimizing the contrastive loss $\mathcal{L}_{cr}$. 

\paragraph{Model Inversion} \label{sec:inversion} In the previous part, we have bridged the data diversity with the contrastive learning objective, which can be directly optimized to make data more diverse. This section integrates contrastive learning into model inversion to form our final algorithm, namely contrastive model inversion.

As illustrated in Figure \ref{fig:ablation_cr}, our approach is consists of four components: a generator $g(\cdot; \theta_g)$, a teacher network $f_t(\cdot; \theta_t)$, an instance discriminator $h(\cdot; \theta_h)$ and a memory bank $\mathcal{B}$. The discriminator is a simple multi-layer perception as used in \cite{chen2020simple}, which accept the representations in the penultimate layer as well as the global pooling of intermediate features as the input. 

The core idea of CMI is to incrementally synthesize some new samples that can be easily distinguished from the historical ones in the memory bank. Thus, the model inversion process is tackled in a ``case-by-case'' strategy, which means in each timestamp $T$, only one batch of data will be synthesized by the generator. Specifically, at the beginning of timestamp $T$, we re-initialize the generator and iteratively optimize its latent code $z$ as well as the parameters $\theta_g$ for synthesis. In this case, the generator is only responsible for a small part of the data distribution. Compared with the pixel updating strategy used in~\cite{yin2020dreaming} that update different pixels independently, the ``case-by-case'' generator can provide stronger regularization on pixels because they are produced from the shared weights $\theta_g$. During synthesis, random augmentation will be applied on the synthetic images to produce a local view $x$ and a global view $x^+$ for contrastive learning. However, note that a single batch will be insufficient for training discriminators training. Thus we let the history images stored in memory bank $\mathcal{B}$ also participate in the learning process. Now, the objective for contrastive model inversion can be formalized as the following:
\begin{equation}
   \min_{\theta_g, z, h} \left[ \alpha_{cr} \cdot \mathcal{L}_{cr}(g(z; \theta_g) \cup \mathcal{B}, h) + \beta_{inv} \cdot \mathcal{L}_{inv}(g(z;\theta_g)) \right]\label{eqn:cmi_final}
\end{equation}
where $\mathcal{L}_{inv}(\cdot)$ refers to the widely used inversion criterion in Equation \ref{eqn:unified}, which is only applied on the global view of images and the term $\mathcal{L}_{cr}$ refers to the proposed contrastive loss for data diversity. Note that $\mathcal{L}_{cr}$ takes both synthetic batch $g(z; \theta_g)$ and history data from $\mathcal{B}$ into account, where the history will provide useful guidance for the current image synthesis. During contrastive learning, we stop the gradient on global view and only allow backpropagation on local ones as done in \cite{chen2020exploring}. We found that this operation can provide more clear gradient for local pattern synthesis.

The full algorithm of contrastive model inversion is summarized in Alg. \ref{alg:cmi}. The synthetic images stored in memory bank $\mathcal{B}$ will be used for downstream distillation tasks.  

\begin{algorithm}[t]
   \DontPrintSemicolon
   \KwIn{A pretrained teacher $f_t(\cdot; \theta_t)$ }
   \KwOut{Image Bank $\mathcal{B}$}   
   {}

   $\mathcal{B} \leftarrow \emptyset$

   initialize discriminator $h(\cdot; \theta_h)$

   \For{number of batches}
   {
       initialize generator $g(\cdot; \theta_g)$
      
      $z \leftarrow \mathcal{N}(0, 1)$

      \For{number of update iterations}
      {

         $x \leftarrow g(z; \theta_g)$ 
         
         $x_{B} \leftarrow sample(\mathcal{B})$

         $\mathcal{L} \leftarrow \alpha_{cr} \cdot \mathcal{L}_{cr}( x \cup x_B, h) + \beta_{inv} \cdot \mathcal{L}_{inv}(x) $

         $z \leftarrow z - \eta \nabla_{z} \mathcal{L} $ 

         $\theta_g \leftarrow \theta_g - \eta \nabla_{\theta_g} \mathcal{L} $ 

         $\theta_h \leftarrow \theta_h - \eta \nabla_{\theta_h} \mathcal{L} $ 
      }

      $\mathcal{B} \leftarrow \mathcal{B} \cup x$  
   }

   return $\mathcal{B}$
   
   \caption{Contrastive Model Inversion.}\label{alg:cmi}
\end{algorithm}

\begin{figure*}[t]
   \begin{center}
      \includegraphics[width=0.95\linewidth]{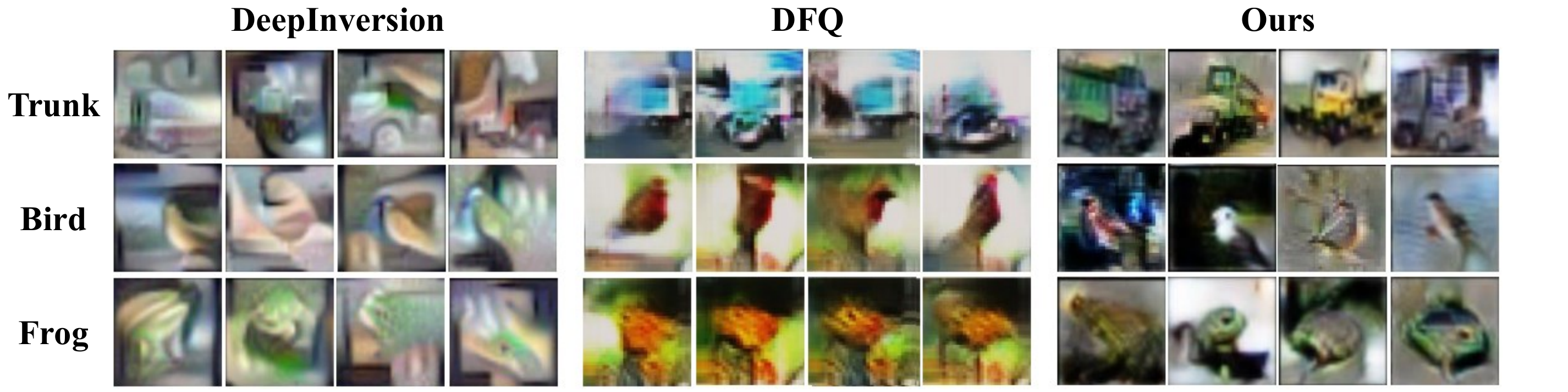}
   \end{center}
   \vspace{-0.5em}
      \caption{Inverted data from a pre-trained ResNet-34 on CIFAR-10. DeepInversion adopts pixel updating for image synthesis, while DFQ uses a generator for generation. Our approach can achieve better visual quality and diversity than the baselines. Besides, the background, style, and pose in our results are very plausible.} \label{fig:visual_quality}
     \vspace{-0em}
\end{figure*}

\begin{table*}[t]
    \centering
    \begin{tabular}{c c c c c c c c c c c}
       \toprule
       \bf \multirow{2}{*}{Dataset} & \bf \multirow{2}{*}{Teacher}  & \bf \multirow{2}{*}{Student}   & \multicolumn{8}{c}{\bf Accuracy}  \\
       \cmidrule(r){4-11} 
       & & & \bf T. & \bf S. & \bf DAFL & \bf ZSKT & \bf ADI & \bf DFQ & \bf LS-GDFD & \bf Ours \\
       \hline 
            \multirow{5}{*}{CIFAR-10} 
            & resnet-34 & resnet-18 & 95.70 & 95.20  & 92.22 & 93.32* & 93.26 & 94.61 & \bf 95.02 & 94.84 \\ 
            & vgg-11    & resnet-18 & 92.25 & 95.20 & 81.10* & 89.46* & 90.36 & 90.84 & N/A & \bf 91.13 \\
            & wrn-40-2  & wrn-16-1 & 94.87 & 91.12 & 65.71* & 83.74 & 83.04* & 86.14 & N/A & \bf 90.01 \\
            & wrn-40-2  & wrn-40-1 & 94.87 & 93.94 & 81.33* & 86.07 & 86.85* & 91.69 & N/A & \bf 92.78 \\ 
            & wrn-40-2  & wrn-16-2 & 94.87 & 93.95 & 81.55* & 89.66 & 89.72* & 92.01 & N/A & \bf 92.52 \\ %
       \hline
            \multirow{5}{*}{CIFAR-100}  
            & resnet-34 & resnet-18 & 78.05 & 77.10 & 74.47 & 67.74* & 61.32* & 77.01 & 77.02 & \bf 77.04 \\ 
            & vgg-11    & resnet-18 & 71.32 & 77.10  & 57.29* & 34.72* & 54.13* & 68.32* & N/A & \bf 70.56 \\ 
            & wrn-40-2  & wrn-16-1 & 75.83 & 65.31 & 22.50* & 30.15* & 53.77* & 54.77* & N/A & \bf 57.91\\ 
            & wrn-40-2  & wrn-40-1 & 75.83 & 72.19 & 34.66* & 29.73* & 61.33* & 61.92* & N/A & \bf 68.88\\
            & wrn-40-2  & wrn-16-2 & 75.83 & 73.56 & 40.00* & 28.44* & 61.34* & 59.01* & N/A & \bf 68.75\\ 
        \hline
            \multirow{1}{*}{Tiny-ImageNet}  
            & resnet-34 & resnet-18 & 66.44 & 64.87 & N/A & N/A & N/A  & 63.73 & N/A & \bf 64.01 \\
        \hline         
    \end{tabular}
    \caption{A benchmark for data-free knowledge distillation on CIFAR-10, CIFAR-100, and Tiny-ImageNet. Method $T$ and $S$ refers to scratch training on labeled data. The results marked by ``*'' come from our re-implementations.} \label{tbl:kd_benchmarks}
 \end{table*}

\subsection{Decision Adversarial Distillation}

With the inverted data $\mathcal{X}$, it is easy to train the student with the Kullback–Leibler divergence. However, the synthetic maybe not optimal for knowledge transfer, where some important patterns are missing. Adversarial distillation is a popular technique to improve student performance, which incorporates the student into image synthesis and maximizes the disagreement between teacher and student using Eqn. \ref{eqn:adv}. However, large disagreement may not always correspond to valuable samples because they can be just some outliers. In this work, we pay more attention to those boundary samples and introduce a decision adversarial loss: 
\begin{equation}
   \begin{split}
      \mathcal{L}_{d-adv} = - K&L(f_t(x)  /  \tau \| f_s(x) / \tau) \cdot \\ 
       & \mathbb{1}\{\arg \max f_t(x) = \arg \max f_s(x)\}\
   \end{split}
   \label{eqn:adv}
\end{equation}
The function $\mathbb{1}\{\cdot\}$ is an indicator to enable adversarial learning when the teacher and student produce the same prediction on $x$, otherwise disable it. Unlike the unbounded loss term in Eqn. \ref{eqn:adv}, our decision adversarial loss will make $x$ close to the decision boundary, which can provide more information about teacher networks.

\section{Experiments}\label{sec:exp}

\subsection{Settings}

\paragraph{Models and Datasets} In this work, we study the effectiveness of CMI on several network architectures, including resnet~\cite{he2015deep}, vgg~\cite{simonyan2014very} and wide resnet~\cite{zagoruyko2016wide}. Three standard classification datasets, CIFAR-10, CIFAR100~\cite{krizhevsky2009learning} and Tiny-ImageNet~\cite{le2015tiny} are utilized to benchmarks existing data-free KD approaches. CIFAR-10 and CIFAR-100 both contain 50,000 samples for training and 10,000 samples for validation, whose resolution is $32\times 32$. Tiny-ImageNet consists of 100,000 training images of resolution $64\times 64$, as well as 10,000 test images, which is more difficult than CIFAR. 

\paragraph{Network Training} In our experiments, all teacher models are trained on labeled datasets, while student models are trained with synthetic data inverted from teachers. We use the Adam Optimizer with 1e-3 learning rate to update the generator and an SGD optimizer with 0.9 momentum and 0.1 learning rate for student training. More details about the network training can be found in supplementary materials. 

\subsection{Benchmarks on Knowledge Distillation}

Table \ref{tbl:kd_benchmarks} provides a benchmark for existing data-free knowledge distillation methods. $T.$ and $S.$ refers to the scratch accuracy of teachers and students on the original training data. We compare our approach with the following baselines: DAFL~\cite{chen2019data}, ZSKT~\cite{micaelli2019zero}, ADI~\cite{yin2020dreaming}, DFQ~\cite{choi2020data} and LS-GDFD~\cite{luo2020large}. They all follow the unified framework discussed in Sec. \ref{sec:preliminary}. Our approach and ADI both adopt a case-by-case strategy~\cite{yin2020dreaming} for model inversion, while the other baselines tackle it as a generative problem. The main difference between our approach and baselines methods lies in the usage of contrastive learning, i.e., the additional loss term $\mathcal{L}_{cr}$. Results show that our approach achieves superior or comparable performance compared with most baseline methods. Besides, note that the wrn-40-2 contains about $10\times$ fewer parameters than the resnet-34 models, where less prior information about training data is stored. In this case, CMI can still achieve satisfying performance compared with baseline methods. 

In Fig.~\ref{fig:visual_quality}, we further visualize the synthetic data from DeepInversion (ADI without adversarial distillation), DFQ, and our approach. The results show that our approach is able to synthesize much more plausible and diverse data. Both DeepInversion and DFQ suffer from model collapse problems. For example, the synthesized frogs in DeepInversion are similar in color and background. It is worth noting that the generator can usually produce better visual quality than pixel update, although it may suffer from more severe mode collapse, as shown in the images from DFQ. In contrast, our approach is able to recover more diverse and plausible results, with better local details such as eyes and textures.

\definecolor{mygreen}{rgb}{0.1,0.8,0.1}
\def\scoreup#1{$(\color{mygreen} \uparrow #1)$}
\def\scoredown#1{$(\color{red} \downarrow #1$)}

\subsection{Ablation Study}


\paragraph{Module Cut-off} Our ablation study begins by investigating the contribution of different module introduced in CMI, including the contrastive loss and the generator described in Sec.\ref{sec:inversion}, as well as the decision adversarial loss. This experiment is organized in a ``cut-off'' schema, where each module will be separately turn off to see its effectiveness. As shown in table \ref{tbl:ablation}, turning off both the contrast loss and generator will lead to poor performance, but it seems cut-off the generator may lead to more severe degradation. We would like to argue that this is reasonable. As mentioned in Section \ref{sec:inversion}, a generator is used to regularize pixels of synthetic data, and a crop operation is applied to construct local and global views. In this case, if we replace the generator with pixel update as used in \cite{yin2020dreaming}, then only the pixels inside the crop region will be updated, leading to an imbalanced optimization. In our method, decision adversarial distillation is also an important technique for knowledge distillation. However, our method without adversarial distillation can still achieve the best performance (87.71) than the state-of-the-art (DFQ, 86.14 on CIFAR-10). 

\begin{table}[t]
   \centering
   \begin{tabular}{l c c c}
   \toprule
      Method & CIFAR-10 & CIFAR-100 \\
      \hline
      CMI & \bf 90.01 & \bf 57.91 \\
      w/o Contrast  & 85.97 & 56.41 \\
      w/o Generator & 83.65 & 54.15 \\
      w/o Decision Adv.   & 87.71 & 56.88 \\
   \hline 
   \end{tabular} 
   \caption{Ablation Study by cutting of different modules. We find a prominent degradation when we cut off the instance classification module, which demonstrate its effectiveness in data-free knowledge distillation.} \label{tbl:ablation}
\end{table} 

\begin{figure}[t]
   \begin{center}
      \includegraphics[width=8cm]{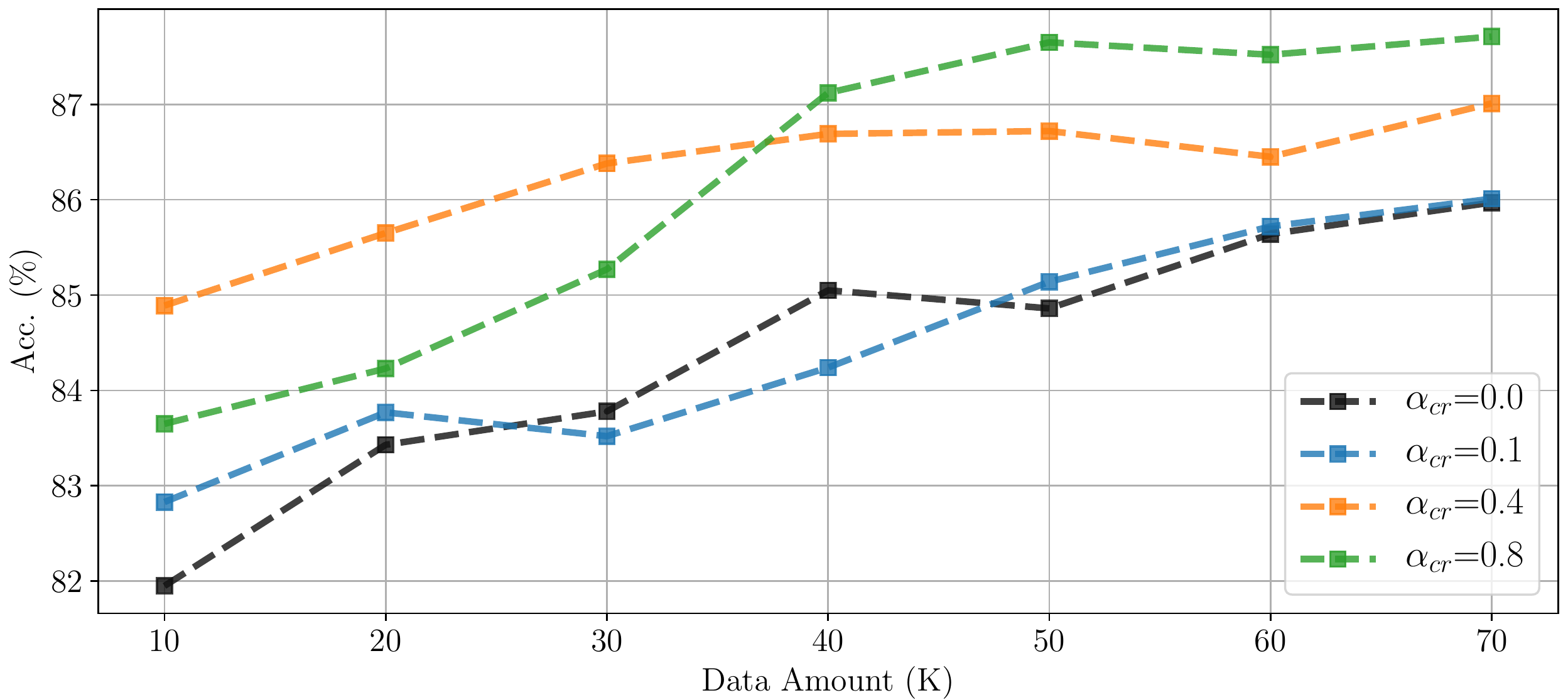}
   \end{center} 
   \vspace{-1mm}
      \caption{We study the influcne of different $\alpha_{cr}$, which controls the contrast level of synthetic data. The results is obtained without adversarial distillation.} \label{fig:ablation_cr}
\end{figure}

\paragraph{Influence of Contrast} In our approach, we can control the data diversity by changing the balance term $\alpha_{cr}$. In this experiment, we take a further investigation on the influence of different $\alpha_{cr}$. Fig. \ref{fig:ablation_cr} illustrates the relation between student performance and $\alpha_{cr}$ with different data amount in both high contrast and low contrast settings. We find that an increasing $\alpha_{cr}$ can bring significant benefits for student learning. In order to investigate how $\alpha_{cr}$ affects the data distribution, we calculate the Frechet Inception Distance (FID) for the synthetic data, which is widely used to evaluate the generation quality in GANs. As shown in table \ref{tbl:fid}, the FID score is estimated on three layers, corresponding to different feature level. The 1-st pooling layer contains more low-level features, while the last pooling layer mainly focuses on high-level semantics. There are two observation in this experiment: first, increasing $\alpha_{cr}$ can indeed achieve a better FID score, i.e., a better distribution; Second, we find significant FID improvements in the shallow layer, which means our approach can effectively improve the synthesis of low-level details. 

\begin{table}[t]
   \centering
   \begin{tabular}{l c c c}
   \toprule
      Feature position & 1-st Pool & 2-nd Pool & final Pool \\
      \hline
      WGAN-GP & N/A & N/A & 29.3 \\
      \hline
      ADI & 2.021 & 17.28 & 84.69\\
      Ours ($\alpha_{cr}$=0.0) & 0.713 & 4.962 & 77.49 \\
      Ours ($\alpha_{cr}$=0.4) & 0.177 & 2.176 & \textbf{62.12} \\
      Ours ($\alpha_{cr}$=0.8) & \textbf{0.140} & \textbf{1.776} & 62.63 \\ 
   \hline 
   \end{tabular} 
   \caption{Fréchet Inception Distance (FID, lower is better) score of synthetic data on CIFAR-10. The FID metric is estimated in both shallow layers and deep layers to assess the synthesis quality from different feature level. } \label{tbl:fid}
\end{table} 

\paragraph{Design of Instance discriminator} Another topic in ablation study lies in the design of an instance classifier. We consider three design of instance discriminator as shown in table \ref{tbl:mlp}. First, we remove the discriminator and directly estimate the contrast in the feature space of the teacher model. This operation would be problematic because this space is not explicitly trained for similarity metric as described in Sec. \ref{sec:diversity}. The second design is a linear discriminator, which is used to select a subset feature for contrastive learning. We compare it with a non-linear one used in our approach. We find that a stronger instance discriminator will be helpful for the student learning.

\begin{table}[t]
    \centering
   \begin{tabular}{l c c c}
   \toprule
      Method & CIFAR-10 & CIFAR-100 \\
      \hline
      None & 88.15 & 56.97 \\
      Linear & 88.57 & 57.01 \\
      Non-Linear & \bf 90.01 & \bf 57.91 \\
   \hline 
   \end{tabular}
   \caption{We study different design of instance Classifier on CIFAR-10 and CIFAR-100 } \label{tbl:mlp}
\end{table} 

\section{Conclusion}

In this work, we propose a novel model inversion approach to guarantee the diversity of synthetic data, which can bring significant benefits for downstream distillation tasks. Extensive experiments are conduct in this paper not only to show the effectiveness of our approach but also explore how it works. However, model inversion is still an interesting topic in the context of knowledge distillation, which may provide some insight on understanding the learned knowledge in a network and how it is transferred to other networks.

\paragraph{Acknowledgement} This work is supported by National Natural Science Foundation of China (U20B2066, 61976186), Key Research and Development Program of Zhejiang Province (2018C01004),  the Fundamental Research Funds for the Central
Universities (2021FZZX001-23) and Alibaba-Zhejiang University Joint Research Institute of Frontier Technologies.


\clearpage

\bibliographystyle{named}
\bibliography{cmi}

\end{document}